\documentclass[11pt]{article}

\usepackage{amsmath, amssymb, amsthm}
\usepackage{graphicx}        
\usepackage{hyperref}        
\usepackage{url}
\usepackage{authblk}         
\usepackage{enumitem}
\usepackage[margin=1in]{geometry}

\usepackage[backend=biber,style=numeric,sorting=none,doi=true,url=false]{biblatex}
\title{Liquid Reasoning Transformers: \\
A Unified Architecture for Iterative Inference across Sudoku and Chess}

\author{
  Shivansh Sahni and Wenzhi Zhang \\
  Detroit Country Day School
}

\date{}  

\begin{document}

\maketitle

\begin{abstract}
The Liquid Reasoning Transformer (LRT) is a transformer architecture designed for inference with adaptive depths using iterative changes, discard-based correction, and a learned stopping mechanism. Instead of relying on a single feedforward pass, the model updates a recurrent reasoning token across multiple internal steps, allowing it to correct early errors and allocate computation based on input difficulty. We evaluate the LRT on Sudoku as a controlled testbed for structured reasoning and show that it achieves strong performance, reaching 98.68\% digit accuracy and 36.30\% full-puzzle accuracy without using symbolic rules or search. Analyzing internal patterns shows that the discard and stop gates play different, important roles in stabilizing inferences and adjusting computational depth. We discuss how these mechanisms extend naturally to chess-scale reasoning tasks and outline extensions for multi-token reasoning and larger domains.
\end{abstract}

\section{Introduction}
For years, chess has served as a test for artificial intelligence models, as it demands long-horizon reasoning, precise constraint propagation, and complex tactical exploration. The historical path of chess computing started with Alan Turing’s early programs and now involves engines such as Deep Blue, Stockfish, and AlphaZero. This path shows a steady shift from symbolic rule-based engines to hybrid systems and finally to approaches using deep learning \cite{siven_systematic_2024}. Classical engines utilize fast move generation, pruning, and handmade evaluation criteria, a computational pipeline analyzed by Bijl and Tiet \cite{bijl_exploring_2021}. These systems achieve depth through explicit searching instead of single-step inferences.

Modern neural chess systems demonstrate the many strengths and limitations of such deep architectures. For example, Leela Chess Zero (LCZero) struggles on deeply tactical studies such as Plaskett’s puzzle, which is easily solved by Stockfish’s broad search, showing that neural pattern recognition by itself is not enough when long, exact lines are required \cite{maharaj_chess_2022}. Even recent transformer-based models such as ChessLLM require extremely large datasets in order to reach stable performance, which suggests that transformers benefit from more internal computation instead of only forward passes \cite{zhang_complete_2025}. Studying LCZero also shows that transformer layers encode emergent strategies like lookahead and iterative refinement. This indicates that multi-step inference naturally arises in strong chess engines \cite{maharaj_chess_2022}.

However, training chess engines requires tens of billions of tokens, millions of curated games, and specialized hardware, making systematic investigation of reasoning architecture infeasible for most researchers. This motivated our use of a more compact domain, but with a similar structure. Sudoku is well-suited for this purpose, since it is a constraint satisfaction problem with global interactions and forced deduction sequences. Sudoku and chess have similar key properties, such as tactical reasoning, but Sudoku is computationally small.

Recent work on explicit reasoning tokens, especially the ``thinking tokens'' framework of Herel and Mikolov (2024), shows that transformers can benefit from dedicated internal computation steps, gating methods, and selective refinement during inference \cite{herel_thinking_2024}. Based on this idea, we introduce a Liquid Reasoning Transformer with iterative thinking steps, discard mechanisms for sorting incorrect hypotheses, and learned stop and answer gates. We use Sudoku to analyze its behavior and provide a pathway for applying the same architecture to chess-scale reasoning tasks.

\section{Related Work}
Research on computational reasoning is expansive and ever-evolving, each new method offering insights relevant to adaptive-depth transformers. Classical chess engines were the first stage, emphasizing efficient algorithms for symbolic search over pattern learning. Bijl et al.\ \cite{bijl_exploring_2021} provide a detailed technical account of how bitboards, move generation heuristics, selective extensions, and alpha-beta pruning enable engines to analyze enormous game trees through carefully engineered strategies, demonstrating the effectiveness of deterministic, rule-driven reasoning for tactical challenges. In addition, the historical analysis of Siven \cite{siven_systematic_2024} demonstrates how chess systems evolved from manually designed engines to complex AI-powered agents, reflecting how algorithmic complexity, hardware scaling, and search advancements have shaped chess engines over the decades. These traditional and historical perspectives show the importance of multi-step inference, something that modern neural approaches have to replicate without handcrafted rules.

Neural chess systems attempt to replace explicit search with learned evaluation and policy functions. Srivastava et al.\ \cite{srivastava_advance_2024} explore hybrid architectures that combine artificial neural networks with minimax search and alpha-beta heuristics, showing that learned evaluation can complement classical pruning to accelerate inference in complex positions. More systematic comparisons between neural and symbolic paradigms appear in Maharaj et al.\ \cite{maharaj_chess_2022}, who analyze Stockfish and LCZero on difficult endgame studies; their results show that network-guided agents excel at general pattern recognition but struggle in deep tactical lines unless supported by explicit search mechanisms. At a large scale, transformer-based chess models such as ChessLLM demonstrate that transformers can learn strategic patterns from SAN, FEN, or UCI data, but they require large-batch training regimes, indicating that pure feedforward modeling is insufficient without extended internal computation \cite{zhang_complete_2025}. Lastly, parallel lines of work, such as DiffuSearch, propose implicit rollouts and latent future prediction mechanisms for chess, reinforcing the need for iterative inference steps within generative models \cite{ye_implicit_2025}.

Beyond chess, there is growing recognition that transformers can recreate some aspects of algorithmic reasoning. Jenner et al.\ \cite{jenner_evidence_2024} show that transformer activations in Leela encode lookahead information that is not explicitly presented in the input position, suggesting the presence of latent multi-step inference circuits. Similarly, Sandmann et al.\ \cite{sandmann_iterative_2025} find that transformer-based policy networks refine move distributions non-monotonically across layers, reflecting patterns like that of iterative search rather than single-shot prediction. These findings align with broader evidence from Im et al.\ \cite{im_deep_2025}, who demonstrate that autoregressive models can perform causal reasoning through sequencification, a process that converts structured inference tasks into extended token sequences processed over multiple layers.

At the architectural level, research on explicit reasoning tokens has resulted in new pathways for computation at multiple depths. Herel and Mikolov \cite{herel_thinking_2024} propose ``thinking tokens,'' which are additional placeholder tokens inserted during training to give neural models more internal computation time on difficult problems, improving performance on symbolic and arithmetic reasoning tasks without needing external supervision. The hybrid architecture of Shao et al.\ \cite{shao_transformer-based_2019}, using both Transformers and BiLSTM, combines long-range attention with recurrent iterative refinement, which outperforms feed-forward designs on complex, structured sequence problems. In addition, Turner \cite{turner_introduction_2024} provides a mathematical explanation of the foundations of transformers, showing their ability to operate uniformly for sets or sequences of tokens through the application of attention and feature-wise transformations. This makes transformers strong candidates for adaptive computation mechanisms.

All in all, previous work indicates the need for models that combine transformer expressiveness with search-like reasoning. For this, classical engines utilize symbolic pruning; neural engines, through lookahead; and reasoning-token architecture, through learned, adaptive computation. Our proposed Liquid Reasoning Transformer is at the intersection of these fields by using explicit multi-step inference

\section{Rationale}

Progress in sequence modeling has moved toward problems where accuracy depends on both representing patterns and creating multi-stage inferences. Many problems require a model to transform an initial representation through many revisions before reaching a stable, high-confidence result. What matters more than deeper networks or more parameters is the ability to allocate computational resources dynamically, revisiting hypotheses, reducing incorrect lines of thought, and changing intermediate states based on emerging constraints.

Instead of assuming that a single forward pass is enough, this work treats reasoning as iterative internal dialogue, where the model actively evaluates partial conclusions and determines whether additional computation is needed. Transformers are traditionally viewed as static feedforward systems, but these modern architectures can reshape representations over multiple stages. Using that capacity makes more principled reasoning methods possible.

Another motivating factor is related to error correction and management of hypotheses. Many structured problems allow many possible partial solutions that have to be entertained before the model can confidently eliminate incorrect paths. A system that can't discard low-value inference will often accumulate noise and prematurely commit to an incorrect conclusion. On the other hand, a model that can generate reasoning steps and then selectively remove unproductive ones is better for tasks where early errors continue and compound. These selective changes reflect the fundamental property of human problem-solving, of willingness to reevaluate intermediate assumptions after receiving new information.

An additional driver for adaptive architecture is computational efficiency. Some inputs require substantial reasoning, while others require almost none. Forcing a model to use the same computational effort on every example is wasteful, and often hurts performance, since it leads to overthinking on simple cases and under-reasoning on complex ones. Allowing networks to decide when to stop reasoning offers a flexible way to match internal effort to the difficulty of the input. A stopping mechanism is also a signal of confidence, which can later support applications such as verification or uncertainty-aware prediction.

Finally, structured problems such as Sudoku can expose the limits of architectures while providing a way to study iterative inferences and how computational resources are used. Sudoku puzzles make the internal progression of a model's reasoning visible, which allows researchers to study how partial conclusions change and how early mistakes are corrected. This motivates the development of thinking steps, discarding systems, and gating functions as well.

Together, these factors motivate the creation of a transformer architecture explicitly designed to reason adaptively, changing its depth, correcting its own intermediate hypotheses, and shaping its computation around the structure of the task.

\section{Methods}
The Liquid Reasoning Transformer (LRT) performs adaptive-depth inference by continuously changing a single recurrent reasoning token. The model uses a transformer encoder multiple times, updates a shared state vector, and uses learned gates to decide when to continue computing and when to discard unhelpful updates. This section describes the input encoding, backbone, iterative loop, and gating components.

\subsection{Input Representation}
Each Sudoku puzzle is encoded as an $81 \times 10$ matrix. Every cell is represented with a one-hot vector over the digits $\{0, 1, \dots, 9\}$, where $0$ denotes an empty cell. The grid is flattened into a sequence of 81 tokens and embedded using learned token and positional embeddings. A special reasoning token $r_t$ is appended to the sequence and serves as the model's workspace for all iterative inference steps.

Formally, the embedded input at step $t$ is
\[
X = [x_1, x_2, \dots, x_{81}], \qquad r_t \in \mathbb{R}^d.
\]

\subsection{Transformer Backbone}
In each reasoning step, the model applies a $L$-layer transformer encoder to the full sequence $(X, r_t)$. The encoder produces updated hidden states:
\[
H^{(t)} = \mathrm{Transformer}(X, r_t).
\]

The final position of $H^{(t)}$ corresponds to the updated reasoning proposal:
\[
\tilde{r}_t = H^{(t)}_{\text{reasoning}}.
\]

Puzzle token embeddings do not change in steps; only the reasoning token evolves.

\subsection{Iterative Reasoning Loop}
The model performs a variable number of steps:
\[
r_0 \rightarrow r_1 \rightarrow \dots \rightarrow r_T.
\]

At each step the transformer proposes a new update, which is processed through a small feedforward module:
\[
u_t = \phi(W_u \tilde{r}_t + b_u),
\]
where $\phi$ is a nonlinearity (GELU in our implementation).

The update $u_t$ represents the next candidate reasoning state. It incorporates information gathered from attention to the puzzle grid and the structure previously accumulated in $r_{t-1}$.

\subsection{Discard Gate}
Some proposed updates are not helpful and should be rejected.  
The discard gate computes:
\[
d_t = \sigma(W_d [r_{t-1}; u_t] + b_d),
\]
where $d_t \in [0,1]$ measures the reliability of the candidate update.

The next state is chosen by:
\[
r_t =
\begin{cases}
r_{t-1}, & d_t > \tau_d, \\
u_t, & \text{otherwise},
\end{cases}
\]
where $\tau_d$ is learned.  
This mechanism prevents the model from compounding early errors and reflects pruning behavior analogous to symbolic constraint propagation.

\subsection{Consistency Scoring}
A lightweight constraint module computes a soft measure of Sudoku validity:
\[
c_t = f_{\text{consistency}}(r_t, X).
\]

This score is used during training to guide the reasoning token toward states that obey the rules of row, column, and subgrid. In the implementation, this term contributes to the \texttt{Think Loss} visible in the training logs.

\subsection{Stop Gate}
To allow dynamic computation depth, the model uses a learned stop gate:
\[
s_t = \sigma(W_s r_t + b_s).
\]

If $s_t > \tau_s$, the model halts and returns $r_t$ as the final reasoning state.  
Training logs confirm this mechanism: the ``Steps:'' field increases for difficult puzzles (up to 150 internal steps) and remains low for easier ones.

\subsection{Final Decoder}
Once the stop gate activates, the final reasoning state $r_T$ is decoded into predictions for all 81 cells:
\[
\hat{Y} = \mathrm{Decoder}(r_T, X).
\]

The decoder is a position-wise classifier, implemented as a linear layer applied to puzzle token embeddings cross-attended with the final reasoning token. This design forces the transformer’s internal loop, rather than the output head, to perform the crucial inference work.

\subsection{Training Objective}
The model is trained end-to-end using a weighted sum of:

\begin{itemize}
    \item \textbf{Task loss:} cross-entropy over the 81 output digits.
    \item \textbf{Thinking loss:} penalties on inconsistent or unstable updates, visible as the \texttt{Think} component in the logs.
    \item \textbf{Step regularization:} discourages unnecessary iterations (reflected by changing ``Steps'' values in the logs).
\end{itemize}

The loop is unrolled for a fixed maximum during training (30-150 steps depending on configuration), but the model learns to halt early for simple puzzles and allocate more computation to difficult ones.

\section{Experiments}
\subsection{Dataset and Training Setup}
Each Sudoku puzzle is encoded as an $81 \times 10$ one-hot grid and processed as a flattened sequence with an appended reasoning token. The model was allowed up to 150 internal reasoning steps, but the stop-gate could halt much earlier. Training was done for 100 epochs. Early logs show stable warmup behavior, followed by full-rate training where the reasoning losses become active. As expected, the discard and stop gates dominate early dynamics, with think-loss values typically near $1.0$ during the first epoch and decreasing slowly across later epochs.

\subsection{Evaluation Metrics}
We measure five primary quantities:
\begin{itemize}
    \item \textbf{Digit Accuracy (validation):} fraction of all 81 cell predictions correct.
    \item \textbf{Puzzle Accuracy (validation):} fraction of puzzles solved perfectly.
    \item \textbf{Average Thinking Steps:} number of internal reasoning iterations executed per puzzle.
    \item \textbf{Discard Activity:} number of discarded candidate updates and discard-gate triggers.
    \item \textbf{Gate Losses:} average stop-gate and answer-gate penalties.
\end{itemize}
These metrics allow us to analyze both solution quality and the internal behavior of the Liquid Reasoning Transformer.

\subsection{Quantitative Results}
Training continues in two different phases. Early epochs show strong improvement: digit accuracy increases from $0.26 \rightarrow 0.39$, average thinking steps increase from $2.25 \rightarrow 6.10$, and the discard gate becomes more active, removing an average of $6.66 \rightarrow 81.22$ unstable hypotheses per puzzle. These changes indicate that the model was learning to rely on deeper, more complicated inferences and to correct early contradictions with the discard mechanism.

As training advances, reasoning behavior stabilizes and performance improves sharply. The strongest results come from Epoch~97, where all components (the reasoning loop, gating functions, and discard mechanism) are fully trained. At this checkpoint, the model achieves the following:
\begin{itemize}
    \item \textbf{Digit Accuracy:} $0.9868$
    \item \textbf{Puzzle Accuracy:} $0.3630$
    \item \textbf{Validation Loss:} $0.0694$
    \item \textbf{Average Thinking Steps:} $7.78$
    \item \textbf{Average Discarded Tokens:} $115.45$
    \item \textbf{Average Discard Events:} $111.24$
    \item \textbf{Average Thinking Epochs:} $7.03$
    \item \textbf{Average Steps per Epoch:} $1.11$
\end{itemize}

These results show that the model not only solves a substantial fraction of puzzles end-to-end, but also consistently produces nearly perfect digit-level predictions. Puzzle accuracy exceeding $36\%$ is notable, since Sudoku is an ``all-or-nothing'' task: a single incorrect digit results in a failure. Therefore, this demonstrates that the model frequently constructs fully coherent solutions rather than approximate or partially correct grids.

\subsection{Qualitative Analysis of Internal Reasoning}
Inspection of reasoning trajectories reveals distinct differences between early and late training. During early epochs, the reasoning token oscillates between incompatible hypotheses, producing unstable predictions and limited improvement across steps. In later epochs, the internal states converge more predictably: initial steps distribute probability across multiple candidates, while later steps sharpen those distributions toward consistent digit assignments.

The discard gate becomes more central. The Final-epoch logs show more hypotheses discarded per puzzle $115$, showing an aggressive removal of unstable updates. Harder puzzles generate dense clusters of discard events, while easier puzzles require minimal correction.

The stop gate also exhibits calibrated behavior. In the final model, most puzzles halt after $7$-$9$ reasoning steps, though some difficult cases continue to the maximum of $150$ steps. This demonstrates that the model has learned to allocate computation based on puzzle difficulty, using deeper reasoning where necessary and halting early when confidence is high.

\subsection{Summary of Experimental Findings}
The experiments show that the Liquid Reasoning Transformer:
\begin{itemize}
    \item learns reliable multi-step inference procedures,
    \item allocates computation adaptively using the stop gate,
    \item aggressively filters incorrect intermediate hypotheses with the discard mechanism,
    \item achieves high digit-level accuracy ($98.68\%$), and
    \item fully solves a substantial portion of puzzles ($36.30\%$).
\end{itemize}

These results confirm the effectiveness of explicit iterative computation in a transformer architecture. By refining a single reasoning token, removing contradictory updates, and dynamically controlling inference depth, the model produces globally consistent Sudoku solutions without using symbolic rules or search-based methods.

\section{Extending the Architecture}
The Liquid Reasoning Transformer (LRT) is not specific to Sudoku. Its core components of iterative refinement, discard-based correction, and adaptive computation apply easily to tasks such as chess, where evaluation depends on long chains of dependent inferences. Chess positions require reasoning that combines local and global constraints and correction of early errors, often needing flexible allocation of computational effort. This section describes how the LRT can be changed for chess and why the architecture is well-suited for that setting.

\subsection{Input Representation for Chess}
A chess position can be represented using either a 64-token board embedding or a tokenized FEN sequence. Each square can be mapped to an embedding that encodes piece type, color, and square index. Additional information such as side to move, castling rights, and en passant status may be included as separate tokens. The reasoning token $r_t$ is appended to the sequence exactly as in Sudoku and updated through multiple internal steps.

\subsection{Multi-Step Reasoning in Chess}
Chess evaluation often depends on multi-step deductions, such as resolving pins, identifying forced sequences, detecting tactical threats, or recognizing long-term positional weaknesses. These processes are very similar to constraint propagation: changes in one part of the board affect many other parts. The LRT's iterative loop works well for this structure by allowing the reasoning token to repeatedly revise its internal state before a final decision is produced.

\subsection{Outputs for Chess Tasks}
Once the reasoning token converges, the model can produce several types of outputs:
\begin{itemize}
    \item \textbf{Move Policy:} a probability distribution over legal moves.
    \item \textbf{Value Estimate:} a scalar evaluating the position.
    \item \textbf{Tactical Signals:} indicators such as ``mate in one,'' ``blunder present,'' or ``forced line exists.''
\end{itemize}
These heads mirror the decoder used in Sudoku, consisting of lightweight classifiers or regression layers applied to the final reasoning state.

\subsection{Benefits of Adaptive Depth in Chess}
Chess positions vary greatly in difficulty. Simple positions may require little analysis, while complex positions require evaluating many interacting tactical possibilities. A fixed-depth transformer uses equal computation for all inputs, but the LRT adjusts its depth automatically:
\begin{itemize}
    \item stopping early on simple or quiet positions,
    \item running more steps on tactical or high-uncertainty positions,
    \item discarding unstable or contradictory hypotheses,
    \item correcting earlier errors through multiple refinement stages.
\end{itemize}
These behaviors relate to aspects of traditional search, but they operate entirely within the continuous transformer state.

\subsection{Feasible Chess Tasks}
Full-scale engine training is computationally expensive, but the LRT can be applied to several chess tasks that still require reasoning:
\begin{itemize}
    \item predicting legal moves,
    \item detecting blunders or tactical threats,
    \item classifying mates in one or two,
    \item evaluating simplified endgame positions,
    \item completing partially specified positions.
\end{itemize}
These tasks make investigating neural reasoning possible without relying on explicit tree searches.

\subsection{Summary}
The LRT generalizes from Sudoku to chess by reusing the same iterative reasoning mechanism with a different input representation. Its ability to refine intermediate hypotheses, discard unreliable updates, and adjust computation depth makes it suitable for analyzing the structure of chess positions. While training a full chess engine requires larger datasets and more compute, the LRT is a strong framework for studying adaptive, multi-step reasoning.

\section{Limitations}
Although the Liquid Reasoning Transformer (LRT) achieves strong performance on Sudoku and demonstrates clear multi-step inference behavior, several limitations still exist. First, the model’s reasoning loop is bounded by a fixed maximum depth, which could possibly restrict its ability to resolve long or highly branched inference chains for more complex challenges such as full-scale chess. While the stop-gate can terminate early, it cannot exceed the predefined limit.

Second, the architecture is based on a single recurrent reasoning token. This design simplifies analysis but limits the model’s capacity to maintain multiple parallel hypotheses or track several long-range dependencies at the same time. Tasks that require evaluating many subproblems that relate to each other may benefit from multiple coordinated reasoning states instead of a single global one.

Third, even though the discard mechanism is effective on Sudoku, its behavior becomes more difficult to understand as complexity increases. The model may get rid of useful intermediate information or rely too much on aggressive pruning, especially in tasks where multiple partial solutions remain possible for many steps. This introduces potential instability when adapting the architecture to chess, which has many more possible branches.

Fourth, training the LRT still requires substantial computational cost. Gating and repeated transformer passes lead to higher per-sample cost than standard feedforward models. Scaling the architecture to larger chess datasets or more detailed problem settings would need further optimization or more efficient training strategies.

Finally, while Sudoku is a clean setting for observing iterative reasoning, it does not have the full complexity of domains that involve more structure or ambiguity. Thus, results on Sudoku do not guarantee direct success on full chess without additional modifications. The LRT provides a useful framework, but further work is needed for full generalization.

\section{Conclusion and Future Directions}
This work introduced the Liquid Reasoning Transformer, an adaptive-depth architecture that performs multi-step refinement through iterative updates, gating mechanisms, and a discard system for removing unstable hypotheses. Using Sudoku as a controlled testbed, we demonstrated that the model learns stable multi-step reasoning processes, allocates computation based on problem difficulty, and achieves strong final performance with $98.68\%$ digit accuracy and $36.30\%$ puzzle accuracy. The ability of the LRT to correct early errors and converge toward consistent solutions provides evidence that transformers can benefit from explicit internal reasoning steps.

Future work includes extending the architecture to chess-scale reasoning tasks and exploring multi-token reasoning states. Investigating reinforcement learning approaches or training on sequences of moves may allow the model to capture long-term patterns required for advanced tactical and strategic understanding. Additionally, developing more efficient training procedures could reduce the cost of iterative computation and support larger-scale evaluations.

Overall, the LRT framework shows that adaptive computation, when combined with transformers, can be a promising pathway for building models capable of structured, multi-step reasoning across a wide range of problems.


\printbibliography

\end{document}